\documentclass{article}

\usepackage[preprint]{neurips_2024}
\usepackage[utf8]{inputenc}
\usepackage[T1]{fontenc}
\usepackage{hyperref}
\usepackage{url}
\usepackage{tabularx}
\newcolumntype{Y}{>{\centering\arraybackslash}X}
\usepackage{amsmath}
\usepackage{booktabs}
\usepackage{xcolor}
\usepackage{multirow}
\usepackage{float}
\usepackage{array}
\usepackage{graphicx}
\usepackage{placeins}

\title{
Fair and Calibrated Toxicity Detection with Robust Training and Abstention}

\author{%
Mokshit Surana \\
University of Illinois Chicago\\
\texttt{msura4@uic.edu}
}

\begin{document}
\maketitle

\begin{abstract}
Fairness in toxicity classification involves three integrated axes: ranking, calibration, and abstention. Training-time interventions and post-hoc safety mechanisms cannot be evaluated independently because the former determines the efficacy of the latter. We compare Empirical Risk Minimization (ERM), instance-level reweighting, and Group DRO across these axes, combined with temperature scaling, confidence-based abstention, and per-identity threshold optimization. Evaluation uses subgroup AUC, BPSN/BNSP AUC, error gaps, and per-subgroup Expected Calibration Error (ECE) with bootstrap CIs ($n = 1000$).

We report four findings. (1) \textbf{Calibration disparity is a hidden fairness violation.} ERM has near-perfect aggregate calibration ($0.013$) but is significantly miscalibrated on every identity subgroup ($+0.029$ to $+0.134$). (2) \textbf{Training interventions reshape rather than eliminate disparity.} Reweighted ERM improves ranking (BPSN AUC $+0.06$ to $+0.12$) but worsens the calibration-fairness gap by up to $+0.232$. Group DRO eliminates calibration disparity but only by becoming uniformly miscalibrated globally (ECE $0.118$). (3) \textbf{Post-hoc methods inherit training failure modes.} Temperature scaling fails because miscalibration is non-uniform. Confidence-based abstention works under ERM but breaks under DRO, where the risk-coverage curve rises with deferral. (4) \textbf{Abstention itself is unfair.} Confidence-based deferral helps background content far more than identity-mentioning content. We argue that SRAI fairness requires a multi-axis framework: methods equivalent on aggregate ranking differ sharply in failure modes that determine real-world harm.
\end{abstract}

\section{Problem and Goal}

\textbf{Problem Statement.} Toxicity classifiers often learn spurious correlations between identity mentions and toxicity labels. Because hate speech frequently co-occurs with mentions of protected groups, models treat identity terms as signals for toxicity. This causes neutral or anti-racist content to be disproportionately penalized, producing systematic false positives. At a fixed probability threshold, these correlations become disparate moderation outcomes.

\textbf{Integrated Fairness Axes.} Standard evaluations focus on ranking metrics (BPSN/BNSP AUC) and treat calibration or abstention as separate quality concerns. We argue this separation is misleading for three reasons. First, calibration disparity is a fairness violation: if confidence corresponds to different accuracies across groups, threshold-based decisions are inherently biased. Second, abstention efficacy is a fairness property: confidence-based deferral only works if confidence tracks correctness uniformly. Third, post-hoc interventions inherit failure modes from training. Temperature scaling, threshold optimization, and abstention succeed or fail based on how the training method shaped the probability distribution.

\textbf{Mapping to SRAI Principles.} The three axes correspond to core Socially Responsible AI functions. \textbf{Protection} from disparate flagging requires fair ranking and fair calibration. \textbf{Information} (uncertainty reporting) requires per-subgroup calibration parity so that a confidence score remains meaningful across all content. \textbf{Prevention and Mitigation} through deferral requires that abstention works uniformly: if deferral is effective only on background content, the safety net is unfair.

\textbf{Contributions.}
\begin{enumerate}
    \item \textbf{Multi-Method Comparison.} We evaluate ERM, Reweighted ERM, and Group DRO combined with three post-hoc interventions, using paired bootstrap 95\% CIs ($n = 1000$) for all metrics.
    \item \textbf{Calibration as Fairness.} We introduce the calibration-fairness gap (subgroup ECE minus background ECE). We show ERM is significantly miscalibrated on every identity subgroup despite near-perfect aggregate calibration.
    \item \textbf{Non-Dominated Trade-offs.} We characterize training methods as occupying distinct positions: ERM has the worst ranking but cleanest abstention; Reweighted ERM has the best ranking but worsens calibration gaps; Group DRO fixes calibration parity but breaks abstention.
    \item \textbf{Post-Hoc Dependency.} We demonstrate that intervention
efficacy is determined by the training method. Abstention works under
ERM, partially under Reweighted ERM, and fails under Group DRO because
DRO's training decouples confidence from correctness.
\end{enumerate}
\section{Related Work}

\textbf{Bias in toxicity classification.} Dixon et al. \citep{dixon2018measuring} documented disparate error rates in early classifiers, while Borkan et al. \citep{borkan2019nuanced} formalized the BPSN and BNSP AUC metrics and released the Civil Comments dataset. Garg et al. \citep{garg2019counterfactual} established that ranking metrics can mask threshold-level deployment failures, motivating our integrated framework pairing ranking with calibration and tail-distribution analysis.

\textbf{Group robustness.} Sagawa et al. \citep{sagawa2020distributionally} formulated Group DRO for fair learning, though later work suggests reweighting often matches its performance \citep{idrissi2022simple}. We replicate this on Civil Comments and show that DRO and reweighting diverge sharply on calibration fairness, an axis prior comparisons did not measure.

\textbf{Calibration as fairness.} Guo et al. \citep{guo2017calibration} introduced temperature scaling to address neural network miscalibration. Pleiss et al. \citep{pleiss2017calibrationfairness} argue that calibration parity across groups is a vital fairness criterion. Our work uses per-subgroup ECE with bootstrap CIs to connect this parity to downstream abstention efficacy.

\textbf{Selective prediction.} Geifman and El-Yaniv \citep{geifman2017selective} formalized confidence-based abstention via risk--coverage curves. We evaluate this connection per subgroup, showing that abstention efficacy varies systematically across identity groups, making the safety mechanism itself a fairness concern.

\textbf{Hate speech generalization.} Mathew et al. \citep{mathew2021hatexplain} released HateXplain for explainability. We use it for zero-shot transfer to frame cross-dataset performance as a deployment and generalization concern rather than a direct fairness comparison.

\section{Data}

\textbf{Civil Comments.} We use the Jigsaw dataset \citep{borkan2019nuanced} containing 1.8 million comments. Toxicity and identity-mention scores are binarized at $0.5$, yielding an $8\%$ positive rate. We stratified-downsample to $200,000$ examples, split $80/10/10$ into train ($160,000$), validation ($20,000$), and test ($20,000$) using \texttt{random\_state=42}.

\textbf{Group Assignment.} Examples are assigned to group $g = (\text{identity}, y)$ for the first identity mentioned. The test set includes 18,217 background examples (no identity mentioned) and eight identity groups with sufficient support: white (276), muslim (247), gay/lesbian (129), black (146), jewish (83), christian, female, and male. Groups with $n < 50$ (e.g., hindu, atheist) are excluded from reported metrics to ensure stable bootstrap estimates.

\textbf{HateXplain.} For zero-shot transfer, we use HateXplain \citep{mathew2021hatexplain}. Majority-vote labels map \textit{hatespeech} to $y=1$ and others to $y=0$. The test split contains $1,924$ examples ($30.9\%$ toxic). We use this to evaluate generalization across domain gaps between news comments and social media posts.

\section{Methods}

We categorize methods into training-time interventions, which produce distinct models, and post-hoc interventions, which operate on trained outputs. These are coupled: the efficacy of a post-hoc mechanism is determined by the training method.

\textbf{ERM Baseline.} We fine-tune \texttt{distilbert-base-uncased} with cross-entropy for 2 epochs, batch size 16, and a linear learning-rate schedule ($5 \times 10^{-5}$ to 0). ERM serves as the baseline, minimizing average loss uniformly with no subgroup awareness.

\textbf{Reweighted ERM.} We apply per-example weights $w_i = N / (G \cdot n_{g_i})$ based on group frequency. Weights are clipped at 50.0 to prevent rare groups from dominating gradients. This serves as a middle ground between ERM and adaptive DRO.

\textbf{Group DRO.} We implement adaptive per-group weights $q_g$ updated each batch via $q_g \leftarrow q_g \cdot \exp(\eta L_g)$ with $\eta = 0.001$. This minimax objective focuses on the highest-loss group.

\textbf{Temperature Scaling.} We learn a scalar $T \in [0.5, 5.0]$ via validation NLL grid search to produce $p_{\text{cal}} = \mathrm{softmax}(\mathbf{z}/T)$. This corrects uniform miscalibration but cannot fix selective, subgroup-specific errors.

\textbf{Confidence-based Abstention.} We compute confidence as $\max(p(x), 1 - p(x))$. At coverage $c$, we retain the top $c$-fraction of predictions to compute error rates, creating risk-coverage curves. This safety mechanism assumes confidence tracks correctness uniformly across subgroups.

\textbf{Per-identity Threshold Optimization.} We grid search $\tau_g \in [0.1, 0.9]$ on validation data to minimize the absolute error gap between subgroups and background. This uniform-shift correction can only repair bias that manifests as a constant probability offset.

\textbf{Calibration Fairness Gap.} We compute ECE separately on each subgroup and the background. The calibration-fairness gap is $\Delta\text{ECE}(g) = \text{ECE}(g) - \text{ECE}(\text{background})$ using 15 equal-width bins. A gap with a CI excluding zero indicates a fairness violation regardless of ranking performance.

\textbf{Statistical Inference.} For all estimates, we run 1000 paired bootstrap iterations. We report means and 95\% CIs ($2.5/97.5$ percentiles). Differences are significant only if the CI excludes zero.

\section{Evaluation Framework}

Fairness is a multi-axis property. Table \ref{tab:eval_framework} maps these axes to specific metrics and interventions.

\begin{table}[h]
\centering
\caption{Integrated evaluation framework. All results include paired bootstrap 95\% CIs.}
\label{tab:eval_framework}
\small
\begin{tabular}{p{2.8cm}p{2.6cm}p{2.8cm}p{5.4cm}}
\toprule
Axis / Interaction & Metric & Methods & Purpose \\
\midrule
Ranking fairness & Subgroup, BPSN, BNSP AUC & ERM, Reweighted, DRO & Measure toxicity ordering within and across subgroups. \\
\addlinespace
Calibration fairness & Subgroup ECE, ECE gap & ERM, Reweighted, DRO & Check if confidence scores are reliable across different groups. \\
\addlinespace
Tail behavior & \% benign $p > 0.9$ & ERM, Reweighted, DRO & Identify "confident-wrong" errors on identity-mentioning content. \\
\addlinespace
Threshold parity & Error gap at $\tau=0.5$ & ERM, Reweighted, DRO & Evaluate deployment-level error rates before optimization. \\
\midrule
Post-hoc coupling & $T^*$, $\tau_g^*$, Risk at $c$ & ERM, Reweighted, DRO × {T-scaling, abstention, threshold opt.}s & Test if post-hoc fixes can repair the specific errors of each trainer. \\
\addlinespace
Generalization & AUC, ECE, BPSN & ERM (HateXplain) & Probe cross-dataset transfer as a deployment concern. \\
\bottomrule
\end{tabular}
\end{table}
\section{Results}

Results are organized around three fairness axes: ranking, calibration, and
abstention. Section~\ref{sec:erm-baseline} establishes the ERM baseline.
Section~\ref{sec:training-interventions} shows how fairness methods reshape
the axes. Section~\ref{sec:posthoc-inheritance} shows that post-hoc
interventions inherit each training method's failure modes.
Section~\ref{sec:integrated-picture} synthesizes the trade-off.
Section~\ref{sec:qualitative} grounds findings in failure cases.
Section~\ref{sec:hatexplain} addresses zero-shot transfer.

\subsection{ERM baseline: hidden calibration disparity}
\label{sec:erm-baseline}

ERM achieves overall AUC $0.940$, ECE $0.013$, and error rate $5.35\%$ --
aggregate metrics that appear strong. Subgroup decomposition reveals two
hidden disparities.

\textbf{Ranking disparity.} Table~\ref{tab:erm_fairness} shows that white,
black, gay/lesbian, and muslim subgroups exhibit BPSN AUC $\leq 0.825$, well
below the overall AUC of $0.940$. Error gaps reach $+0.199$ for white,
where the subgroup error rate is $\approx 4\times$ the background rate. High
BNSP alongside low BPSN is the signature of identity mentions acting as
toxicity signals.

\begin{table}[h]
\centering
\caption{ERM subgroup fairness. ``n/a'' = fewer than 50 toxic subgroup
examples for stable BNSP estimation.}
\label{tab:erm_fairness}
\small
\begin{tabular}{lrcccc}
\toprule
Identity & $n$ & Sub. AUC & BPSN & BNSP & Error gap \\
\midrule
white       & 276 & 0.763 & 0.780 & 0.948 & $+0.199$ \\
black       & 146 & 0.756 & 0.764 & n/a   & $+0.167$ \\
muslim      & 247 & 0.868 & 0.823 & n/a   & $+0.093$ \\
gay/lesbian & 129 & 0.848 & 0.801 & n/a   & $+0.165$ \\
jewish      &  83 & 0.885 & 0.850 & n/a   & $+0.116$ \\
christian   & 474 & 0.933 & 0.912 & 0.951 & $+0.038$ \\
female      & 566 & 0.902 & 0.880 & 0.954 & $+0.036$ \\
male        & 506 & 0.910 & 0.865 & 0.964 & $+0.051$ \\
\bottomrule
\end{tabular}
\end{table}

\textbf{Calibration disparity.} Despite overall ECE $0.013$, every identity
subgroup has significantly higher ECE than background (Table~\ref{tab:erm_calib_gap},
all CIs exclude zero). The gap reaches $+0.134$ on jewish and $+0.087$ on
gay/lesbian. The model is well calibrated on bulk content but systematically
overconfident on identity-mentioning content. This is a fairness violation
BPSN cannot detect: a prediction of $p = 0.85$ on white-mentioning content
does not correspond to the same accuracy as $p = 0.85$ on background.

\begin{table}[h]
\centering
\caption{ERM calibration-fairness gap. Background ECE $= 0.0099$
($n = 18{,}217$). All gaps significant (CIs exclude zero).}
\label{tab:erm_calib_gap}
\small
\begin{tabular}{lrccc}
\toprule
Identity & $n$ & ECE & Gap & 95\% CI \\
\midrule
white       & 276 & 0.087 & $+0.077$ & $[+0.064, +0.144]$ \\
black       & 146 & 0.095 & $+0.085$ & $[+0.077, +0.180]$ \\
muslim      & 247 & 0.071 & $+0.061$ & $[+0.053, +0.111]$ \\
christian   & 474 & 0.042 & $+0.032$ & $[+0.025, +0.060]$ \\
jewish      &  83 & 0.144 & $+0.134$ & $[+0.096, +0.205]$ \\
female      & 566 & 0.039 & $+0.029$ & $[+0.026, +0.060]$ \\
male        & 506 & 0.043 & $+0.033$ & $[+0.030, +0.066]$ \\
gay/lesbian & 129 & 0.096 & $+0.087$ & $[+0.075, +0.166]$ \\
\bottomrule
\end{tabular}
\end{table}

\subsection{Training-time interventions}
\label{sec:training-interventions}

Table~\ref{tab:overall} shows aggregate metrics for all three methods. Both
fairness methods produce significant AUC drops. DRO's ECE rises $10\times$;
Reweighted ECE rises $3\times$. Aggregate ECE is uninformative without
subgroup decomposition, the methods produce categorically different
calibration distributions.

\begin{table}[h]
\centering
\caption{Aggregate test metrics. AUC drops vs.\ ERM significant for both
methods (Reweighted CI $[-0.019, -0.008]$; DRO CI $[-0.016, -0.007]$).}
\label{tab:overall}
\begin{tabular}{lccc}
\toprule
Metric & ERM & Reweighted & Group DRO \\
\midrule
Overall AUC & 0.9398 & 0.9259 & 0.9284 \\
Overall ECE & 0.0133 & 0.0418 & 0.1344 \\
\bottomrule
\end{tabular}
\end{table}

\textbf{Ranking axis.} Both fairness methods improve BPSN AUC on all eight
identities (Table~\ref{tab:threeway_bpsn}, all CIs exclude zero) with
Reweighted ERM leading on 7 of 8. Both simultaneously reduce BNSP on every
measurable identity, confirming a genuine fairness--accuracy trade-off.
Subgroup AUC is unchanged across methods, showing the methods shift
ranking across groups rather than improving within-group discrimination.

\begin{table}[h]
\centering
\caption{Three-way BPSN/BNSP. Bold = best BPSN. All BPSN gains and
measurable BNSP losses significant (paired bootstrap CIs exclude zero).}
\label{tab:threeway_bpsn}
\small
\begin{tabular}{l|ccc|ccc}
\toprule
 & \multicolumn{3}{c|}{BPSN AUC} & \multicolumn{3}{c}{BNSP AUC} \\
Identity & ERM & Rew. & DRO & ERM & Rew. & DRO \\
\midrule
white       & 0.780 & \textbf{0.884} & 0.866 & 0.948 & 0.830 & 0.855 \\
black       & 0.764 & 0.822 & \textbf{0.839} & n/a & n/a & n/a \\
muslim      & 0.823 & \textbf{0.910} & 0.894 & n/a & n/a & n/a \\
christian   & 0.912 & \textbf{0.972} & 0.938 & 0.951 & 0.831 & 0.903 \\
jewish      & 0.850 & \textbf{0.928} & 0.883 & n/a & n/a & n/a \\
female      & 0.880 & \textbf{0.940} & 0.892 & 0.954 & 0.852 & 0.926 \\
male        & 0.865 & \textbf{0.925} & 0.885 & 0.964 & 0.857 & 0.914 \\
gay/lesbian & 0.801 & \textbf{0.917} & 0.896 & n/a & n/a & n/a \\
\bottomrule
\end{tabular}
\end{table}

\begin{figure}[h]
\centering
\includegraphics[width=0.95\textwidth]{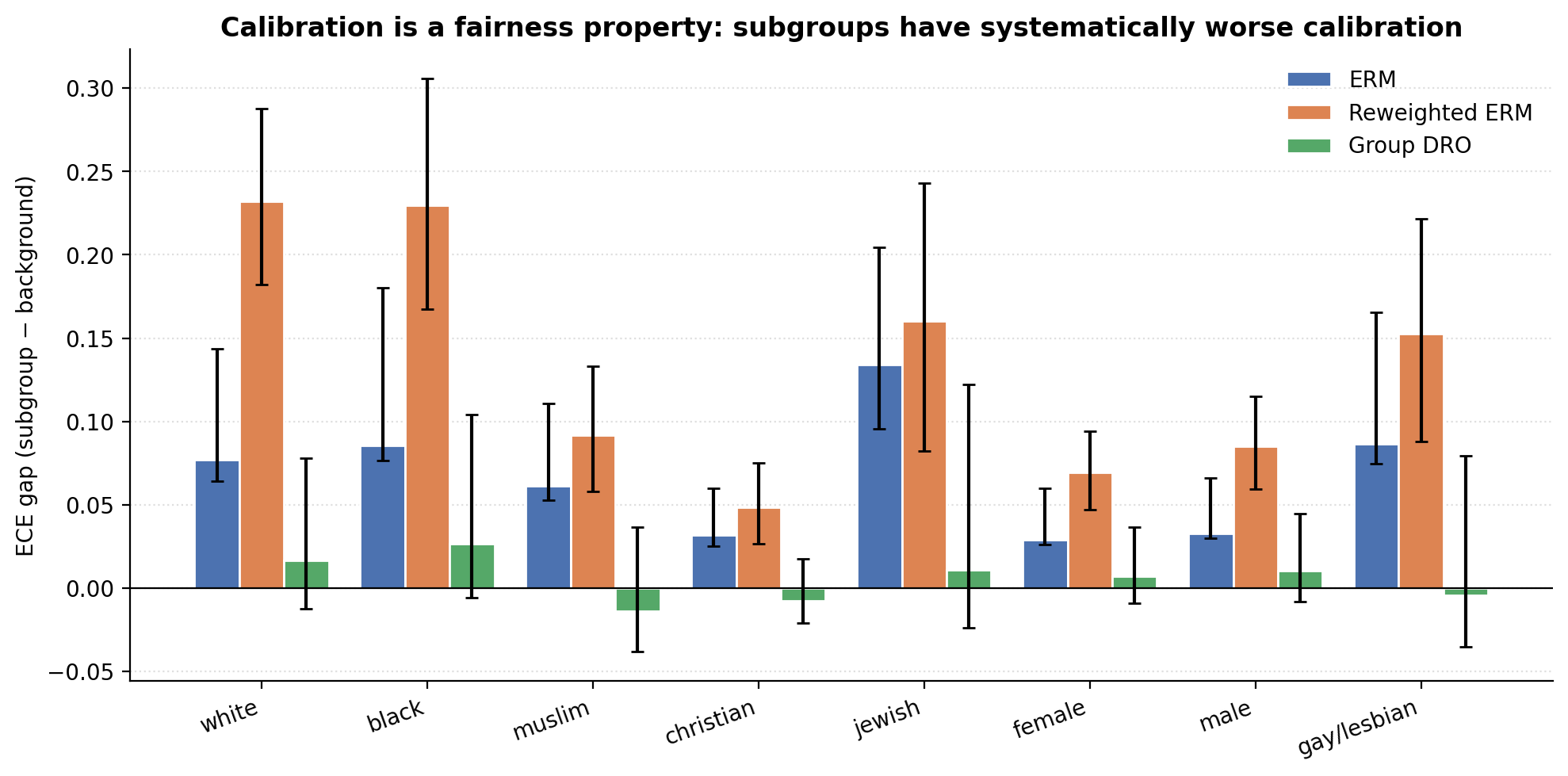}
\caption{Calibration-fairness gap by method. ERM has significant disparity
on all eight identities. Reweighted worsens it substantially. DRO eliminates
it but at the cost of uniform global miscalibration.}
\label{fig:calib_gap}
\end{figure}

\begin{figure}[h]
\centering
\includegraphics[width=\textwidth]{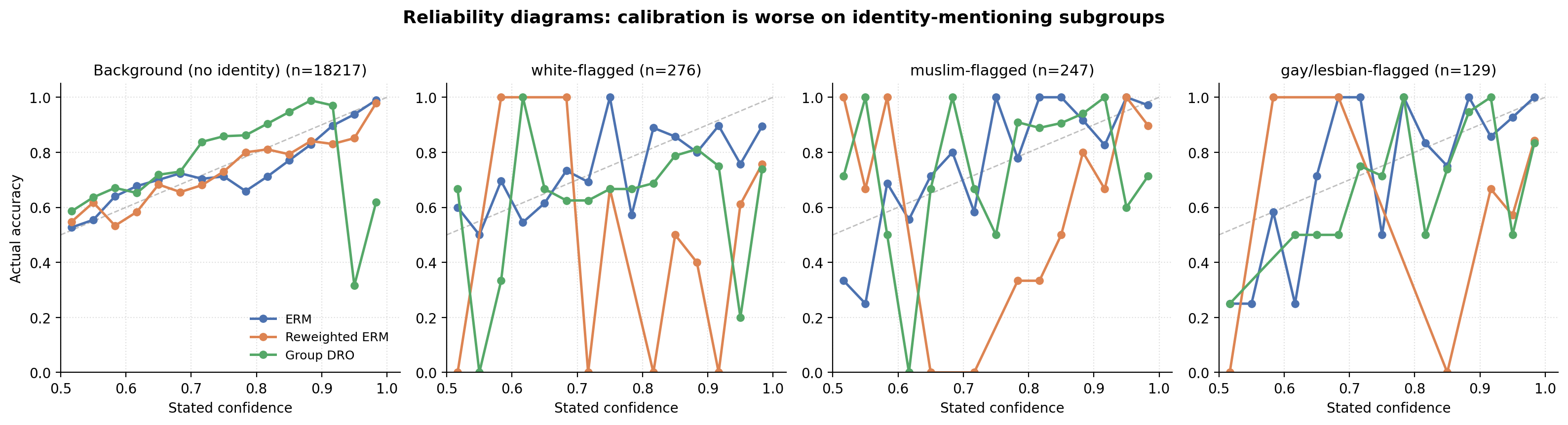}
\caption{Reliability diagrams by subgroup. Background curves hug the
diagonal; identity subgroup curves deviate substantially. Subgroup curves
are noisy due to small $n$; Table~\ref{tab:calib_gap_three_way} is the
reliable summary.}
\label{fig:reliability}
\end{figure}

\textbf{Calibration axis.} Figure~\ref{fig:calib_gap} and
Table~\ref{tab:calib_gap_three_way} show three qualitatively different
calibration profiles. ERM has hidden subgroup disparity (background ECE
$0.010$, but every subgroup significantly miscalibrated). Reweighted ERM
amplifies the disparity: background ECE rises modestly to $0.025$, while
subgroup gaps reach $+0.232$ on white and $+0.230$ on black, approximately
$3\times$ ERM's gaps, the fairness intervention worsened calibration
disparity. Group DRO eliminates subgroup disparity (every gap CI crosses
zero) but only by becoming uniformly miscalibrated everywhere (background
ECE $0.118$). Reliability diagrams (Figure~\ref{fig:reliability}) confirm
these patterns visually.

\begin{table}[h]
\centering
\caption{Calibration-fairness gap (subgroup ECE $-$ background ECE).
\textbf{Bold} = CI excludes zero. Background ECE: ERM $0.010$,
Reweighted $0.025$, DRO $0.118$.}
\label{tab:calib_gap_three_way}
\small
\begin{tabular}{l|ccc}
\toprule
Identity & ERM gap & Reweighted gap & DRO gap \\
\midrule
white       & \textbf{$+0.077$} \,$[0.064, 0.144]$ & \textbf{$+0.232$} \,$[0.182, 0.288]$ & $+0.017$ \,$[-0.012, +0.078]$ \\
black       & \textbf{$+0.085$} \,$[0.077, 0.180]$ & \textbf{$+0.230$} \,$[0.168, 0.306]$ & $+0.027$ \,$[-0.006, +0.104]$ \\
muslim      & \textbf{$+0.061$} \,$[0.053, 0.111]$ & \textbf{$+0.092$} \,$[0.058, 0.133]$ & $-0.014$ \,$[-0.038, +0.037]$ \\
christian   & \textbf{$+0.032$} \,$[0.025, 0.060]$ & \textbf{$+0.049$} \,$[0.027, 0.075]$ & $-0.008$ \,$[-0.021, +0.018]$ \\
jewish      & \textbf{$+0.134$} \,$[0.096, 0.205]$ & \textbf{$+0.160$} \,$[0.082, 0.243]$ & $+0.011$ \,$[-0.024, +0.122]$ \\
female      & \textbf{$+0.029$} \,$[0.026, 0.060]$ & \textbf{$+0.069$} \,$[0.047, 0.094]$ & $+0.007$ \,$[-0.009, +0.037]$ \\
male        & \textbf{$+0.033$} \,$[0.030, 0.066]$ & \textbf{$+0.085$} \,$[0.060, 0.115]$ & $+0.010$ \,$[-0.008, +0.045]$ \\
gay/lesbian & \textbf{$+0.087$} \,$[0.075, 0.166]$ & \textbf{$+0.153$} \,$[0.088, 0.221]$ & $-0.004$ \,$[-0.035, +0.080]$ \\
\bottomrule
\end{tabular}
\end{table}

\begin{figure}[h]
\centering
\includegraphics[width=\textwidth]{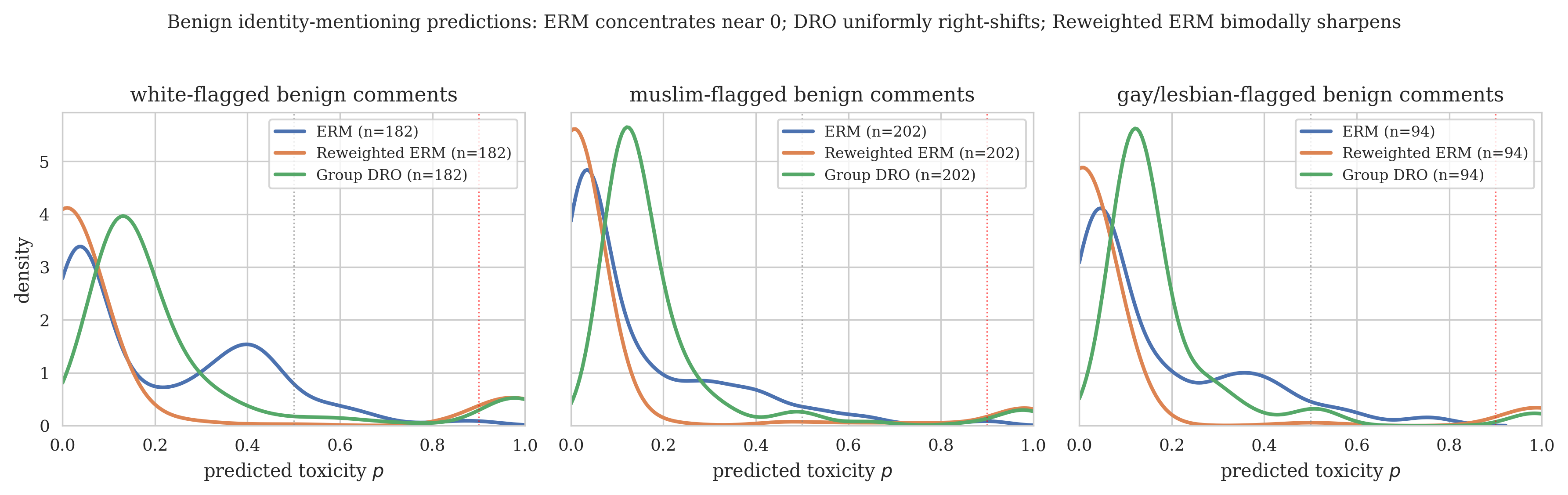}
\caption{Predicted toxicity distributions on benign identity-flagged
comments. ERM concentrates near zero; DRO right-shifts; Reweighted ERM
sharpens toward zero but develops a $p > 0.9$ tail.}
\label{fig:tail}
\end{figure}

\textbf{Tail axis.} Table~\ref{tab:tail} and Figure~\ref{fig:tail} show
that similar BPSN gains arise from categorically different distribution
shifts. DRO uniformly right-shifts all benign predictions (mean $p$:
$0.214 \to 0.253$ for white; $68$--$76\%$ of examples move up). Reweighted
ERM bimodally sharpens: the mean falls ($0.214 \to 0.129$) as most examples
move toward zero, but a small tail is pushed to $p > 0.99$. Critically, ERM
produces zero benign-at-$p > 0.99$ predictions; both fairness methods
produce $1.5$--$4.3\%$. These confidently-wrong predictions would auto-flag
benign identity speech in any pipeline using a high-confidence removal
threshold. BPSN AUC cannot detect this failure mode.

\begin{table}[h]
\centering
\caption{Tail-distribution probe on benign identity-mentioning comments.
Both fairness methods produce $1.5$--$4.3\%$ benign-at-$p>0.99$; ERM
produces none.}
\label{tab:tail}
\small
\begin{tabular}{l|l|rrrr}
\toprule
Identity & Method & mean $p$ & $p{>}0.9$ & $p{>}0.99$ & mean $\Delta$ \\
\midrule
\multirow{3}{*}{white}
 & ERM            & 0.214 & 0.5\%  & 0.0\% & - \\
 & Reweighted     & 0.129 & 10.4\% & \textbf{3.8\%} & $-0.085$ \\
 & DRO            & 0.253 & 9.3\%  & \textbf{3.3\%} & $+0.039$ \\
\midrule
\multirow{3}{*}{muslim}
 & ERM            & 0.147 & 0.5\%  & 0.0\% & - \\
 & Reweighted     & 0.082 & 5.4\%  & \textbf{2.5\%} & $-0.065$ \\
 & DRO            & 0.198 & 4.0\%  & \textbf{1.5\%} & $+0.051$ \\
\midrule
\multirow{3}{*}{gay/lesbian}
 & ERM            & 0.167 & 0.0\%  & 0.0\% & - \\
 & Reweighted     & 0.076 & 6.4\%  & \textbf{4.3\%} & $-0.091$ \\
 & DRO            & 0.189 & 3.2\%  & \textbf{2.1\%} & $+0.022$ \\
\bottomrule
\end{tabular}
\end{table}
\subsection{Post-hoc interventions: inheriting training failure modes}
\label{sec:posthoc-inheritance}

\textbf{Temperature scaling.} Table~\ref{tab:tempscale} shows $T^* = 1.0$
for all three models, not a uniform null result, but three distinct
failures. ERM needs no fix (already calibrated globally). Reweighted ERM
has selective miscalibration concentrated in identity subgroups; a scalar
cannot repair subgroup-specific errors. DRO has structural decoupling of
confidence from correctness; a scalar cannot restore a broken
confidence-correctness relationship. Temperature scaling is a uniform-shift
correction, and none of the training methods produce uniform-shift
miscalibration.

\begin{table}[h]
\centering
\caption{Temperature scaling results. $T^* = 1.0$ for all models;
ECE is unchanged. The failure reason differs by training method.}
\label{tab:tempscale}
\small
\begin{tabular}{lccp{5.5cm}}
\toprule
Method & ECE & $T^*$ & Failure reason \\
\midrule
ERM & 0.013 & 1.0 & Already calibrated globally; subgroup disparity is beyond scalar reach. \\
Reweighted & 0.042 & 1.0 & Non-uniform miscalibration; concentrated in identity subgroups. \\
Group DRO & 0.134 & 1.0 & Structural: confidence-correctness decoupled globally. \\
\bottomrule
\end{tabular}
\end{table}

\textbf{Per-identity threshold optimization.} Table~\ref{tab:errgap} shows
error gaps across methods. On ERM, threshold optimization reduces the white
gap by only $-0.002$ and worsens muslim by $+0.019$. No method achieves
threshold-level parity. The reason mirrors temperature scaling: threshold
optimization corrects uniform-shift bias, but the bias here is tail-heavy.
Most benign comments are correctly scored near zero; a small tail is pushed
to extreme confidence. No single threshold separates confidently-wrong benign
comments from genuinely toxic ones in the same score range.

\begin{table}[h]
\centering
\caption{Error gap (subgroup $-$ background) at $\tau = 0.5$ across methods,
and per-identity threshold optimization on ERM.}
\label{tab:errgap}
\small
\begin{tabular}{lcccc}
\toprule
Identity & ERM & Reweighted & DRO & ERM $+$ $\tau^*$ \\
\midrule
white       & $+0.199$ & $+0.217$ & $+0.205$ & $+0.197$ \\
black       & $+0.167$ & $+0.228$ & $+0.147$ & - \\
muslim      & $+0.093$ & $+0.069$ & $+0.070$ & $+0.112$ \\
christian   & $+0.038$ & $+0.017$ & $+0.012$ & $+0.030$ \\
female      & $+0.036$ & $+0.041$ & $+0.038$ & $+0.047$ \\
male        & $+0.051$ & $+0.059$ & $+0.059$ & $+0.047$ \\
gay/lesbian & $+0.165$ & $+0.126$ & $+0.137$ & - \\
\bottomrule
\end{tabular}
\end{table}

\begin{figure}[H]
\centering
\includegraphics[width=\textwidth]{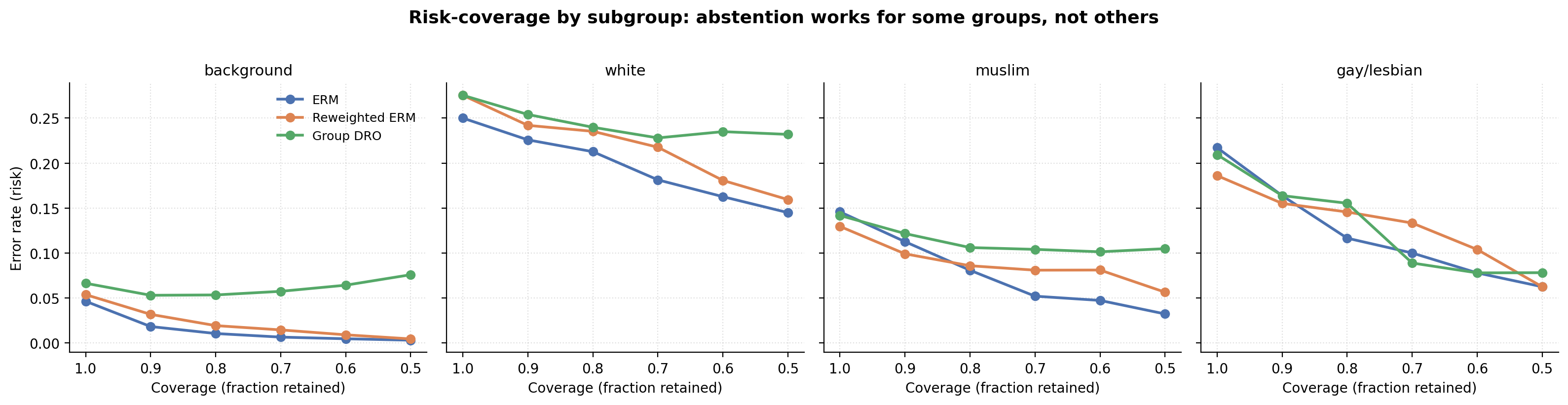}
\caption{Per-subgroup risk-coverage. ERM and Reweighted ERM show clean
abstention on background. White subgroup residual risk stays $\geq 14\%$
across all methods and coverage levels. DRO shows minimal abstention benefit
on any subgroup.}
\label{fig:rc_subgroup}
\end{figure}

\textbf{Abstention.} Figure~\ref{fig:rc_subgroup} shows per-subgroup
risk-coverage curves. ERM's overall risk drops from $5.35\%$ to $0.75\%$
at coverage $0.7$ -- textbook selective prediction behavior. Reweighted
ERM drops smoothly but less steeply. Group DRO's curve rises with
deferral because its ECE of $0.134$ structurally decouples confidence
from correctness: deferring low-confidence predictions does not
preferentially defer wrong ones. \emph{Group DRO and abstention do not
compose.}

Abstention is also unfair across subgroups (Figure~\ref{fig:rc_subgroup}).
On background, ERM drives risk near zero by coverage $0.5$. On the white
subgroup, residual risk persists at $\geq 14\%$ at every coverage level for
all three methods. Muslim and gay/lesbian fall between these extremes.
Abstention provides substantially less safety on identity-mentioning content
because the errors are confidently-wrong rather than uncertain, they survive
the confidence filter. Table~\ref{tab:abstain_erm} shows that muslim, female,
and male gaps close under ERM abstention; the white gap persists. The
training-method choice determines which subgroup receives safe deferral.

\begin{table}[H]
\centering
\caption{ERM abstention error gaps. Muslim, female, male gaps shrink toward
zero by $80\%$ coverage. White gap persists at $\geq 0.11$, signature of
confidently-wrong predictions surviving the confidence filter.}
\label{tab:abstain_erm}
\begin{tabular}{lcrrrr}
\toprule
Coverage & Risk & white & muslim & female & male \\
\midrule
1.0 & 5.35\% & $+0.199$ & $+0.093$ & $+0.036$ & $+0.051$ \\
0.9 & 2.41\% & $+0.131$ & $+0.022$ & $+0.020$ & $+0.009$ \\
0.8 & 1.26\% & $+0.135$ & $+0.008$ & $+0.011$ & $-0.002$ \\
0.7 & 0.75\% & $+0.114$ & $+0.037$ & $+0.001$ & $-0.003$ \\
\bottomrule
\end{tabular}
\end{table}

\subsection{Integrated picture: three axes, no dominant method}
\label{sec:integrated-picture}

Table~\ref{tab:integrated} summarizes each method's position across the three
axes and post-hoc compatibility. No method is Pareto-dominant. ERM has the
worst ranking fairness but cleanest abstention. Reweighted ERM has the best
ranking fairness but worst calibration disparity. Group DRO eliminates
calibration disparity but only by becoming uniformly miscalibrated, breaking
abstention entirely. The choice of training method is a choice of which
fairness axis to prioritize and which post-hoc mechanism to retain.

Finally, zero-shot transfer to HateXplain collapses AUC from $0.940$ to
$0.564$ with BPSN values near $0.5$; we treat this as a
deployment-generalization finding since near-random in-domain performance
makes fairness comparisons uninterpretable.

\begin{table}[H]
\centering
\caption{Three-axis summary. No method dominates across all columns.}
\label{tab:integrated}
\small
\begin{tabularx}{\textwidth}{l|YYY|YY}
\toprule
 & \multicolumn{3}{c|}{Fairness axes} & \multicolumn{2}{c}{Post-hoc} \\
Method & Ranking & Calibration & Tail & T-scaling & Abstention \\
\midrule
ERM        & worst BPSN    & hidden disparity        & no $p{>}0.99$          & not needed       & works \\
Reweighted & best BPSN     & worst disparity         & ${\leq}4.3\%$ at $p{>}0.99$ & cannot fix  & partial \\
Group DRO  & 2nd best BPSN & uniform-bad/no disparity & ${\leq}3.3\%$ at $p{>}0.99$ & cannot fix & broken \\
\bottomrule
\end{tabularx}
\end{table}

\subsection{Qualitative failure cases}
\label{sec:qualitative}

Table~\ref{tab:qualitative} shows benign white-flagged comments where ERM
already predicts $p > 0.6$. All five span all three axes simultaneously:
ranking failure (scored above toxic background), calibration failure (high
stated confidence, zero actual correctness), and tail failure (clustering at
$p \geq 0.9$). Row 14132 is the only case where DRO correctly reduces the
prediction (0.65 to 0.14) while Reweighted worsens it (0.90), showing the
methods do not fail uniformly. These cases explain why ERM's white abstention
gap persists: the errors are high-confidence, so the confidence filter cannot
remove them.

\begin{table}[H]
\centering
\caption{Benign ($y=0$) white-flagged failure cases. \textbf{Bold} = $p>0.99$.}
\label{tab:qualitative}
\small
\begin{tabular}{p{1.8cm}p{7.2cm}rrr}
\toprule
Failure mode & Comment (truncated) & ERM & Rew & DRO \\
\midrule
Anti-racist & ``...what the Nazi scum in Charlottesville did...'' & 0.88 & \textbf{0.99} & \textbf{0.99} \\
Anti-racist & ``Anyone who says such a racies's trash needs to taste trash himself!'' & 0.81 & \textbf{0.99} & \textbf{0.98} \\
Descriptive & ``His skin is white like the Clintons.'' & 0.91 & \textbf{0.99} & \textbf{0.99} \\
Statistical & ``Cops shoot white folks about half as often as black folks.'' & 0.68 & \textbf{0.99} & \textbf{0.98} \\
Political & ``...is an old white guy. The irony is overwhelming.'' & 0.65 & 0.90 & 0.14 \\
\bottomrule
\end{tabular}
\end{table}

\subsection{Limitations}
\label{sec:limitations}

\textbf{Sparse minority groups.} Rare identity groups have $n < 250$ in our
test set (jewish: 83, gay/lesbian: 129, black: 146), producing wide CIs.
Stratified identity oversampling would address this and likely stabilize DRO
weight dynamics; our fix (reducing $\eta$ from $0.01$ to $0.001$) left
background at $83\%$ of group weight.

\textbf{First-match group assignment.} Multi-identity comments contribute to
only one group's training signal and evaluation, potentially understating
intersectional harms.

\textbf{Single model family.} All experiments use DistilBERT-base. The
mechanisms we document likely reflect training-objective properties rather
than architecture, but replication on larger models is needed.

\textbf{ECE binning sensitivity.} We use 15 equal-width bins throughout.
Alternative binning would shift absolute ECE values but not the qualitative
ordering of methods or the significance of calibration-fairness gaps.

\section{Conclusion}

Fairness in toxicity classification involves three integrated axes - ranking,
calibration, and abstention, and training-time interventions determine what
post-hoc mechanisms can repair. We showed this through a bootstrap-validated
comparison of ERM, Reweighted ERM, and Group DRO on Civil Comments, jointly
evaluated with temperature scaling, abstention, and threshold optimization.

ERM is significantly miscalibrated on every identity subgroup despite
near-perfect aggregate calibration, a fairness violation BPSN cannot detect.
Reweighted ERM improves ranking fairness but worsens calibration disparity.
Group DRO eliminates calibration disparity but only by becoming uniformly
miscalibrated, breaking abstention. Each post-hoc intervention fails in ways
determined by the training method: temperature scaling finds $T^* = 1.0$ for
three distinct reasons, threshold optimization cannot repair tail-heavy bias,
and abstention breaks under DRO while remaining unfair on white-flagged
content under all methods.

SRAI fairness evaluation must combine ranking metrics, per-subgroup
calibration gaps with bootstrap CIs, per-subgroup abstention behavior, and
qualitative failure-mode analysis. Methods equivalent on aggregate ranking
differ sharply in failure modes. On this benchmark, Reweighted ERM is the
more defensible practical choice, better calibration, abstention
compatibility, equivalent ranking, but both fairness methods introduce
confidently-wrong predictions on benign identity content that ERM does not,
and no post-hoc intervention repairs this.
\bibliographystyle{plain}

\end{document}